\documentclass{article}
\usepackage{ijcai20}

\usepackage{algorithmic}
\usepackage{amssymb}
\usepackage{color}
\usepackage{multirow}
\usepackage{bbding}
\usepackage{array}
\usepackage{times}

\usepackage{soul}
\usepackage{url}
\usepackage[hidelinks]{hyperref}
\usepackage[utf8]{inputenc}
\usepackage[small]{caption}
\usepackage{graphicx}
\usepackage{amsmath}
\usepackage{booktabs}
\title{Deep Interleaved Network for Image Super-Resolution With Asymmetric Co-Attention}

\author{
Feng Li$^{1,2}$\thanks{Equal Contributions.}\and
Runming Cong$^{1,2}$\footnotemark[1]\and
Huihui Bai$^{1,2}$\thanks{Corresponding Author}\And
Yifan He$^{1,2}$\\
\affiliations
$^1$Institute of Information Science, Beijing Jiaotong University\\
$^2$Beijing Key Laboratory of Advanced Information Science and Network Technology\\
\emails
\{l1feng, rmcong, hhbai, yifanhe\}@bjtu.edu.cn}
\begin{document}

\maketitle

\begin{abstract}
Recently, Convolutional Neural Networks (CNN) based image super-resolution (SR) have shown significant success in the literature. However, these methods are implemented as single-path stream to enrich feature maps from the input for the final prediction, which fail to fully incorporate former low-level features into later high-level features. In this paper, to tackle this problem, we propose a deep interleaved network (DIN) to learn how information at different states should be combined for image SR where shallow information guides deep representative features prediction. Our DIN follows a multi-branch pattern allowing multiple interconnected branches to interleave and fuse at different states. Besides, the asymmetric co-attention (AsyCA) is proposed and attacked to the interleaved nodes to adaptively emphasize informative features from different states and improve the discriminative ability of networks. Extensive experiments demonstrate the superiority of our proposed DIN in comparison with the state-of-the-art SR methods. 
\end{abstract}

\section{Introduction}
Single image super-resolution (SISR), with the goal of recovering a high-resolution (HR) image from its low-resolution (LR) counterpart, is a classical low-level computer vision task and has received much attention. SISR is an ill-posed inverse problem since a multitude of HR images can be degraded to LR one. There are numerous image SR methods that have been proposed to solve such problem including interpolation-based approach~\cite{sparse}, reconstruction-based approach~\cite{edge2010}, and example-based approach~\cite{selfexsr,rfl}.

Recently, inspired by the powerful learning ability of convolutional neural networks (CNN) in computer vision, many CNN based SISR methods~\cite{srcnn,vdsr,lapsrn,edsr,drrn,rdn} have been proposed to learn the end-to-end mapping function from a LR input to its corresponding HR output. Dong~\emph{et al.}~\cite{srcnn} firstly introduce a shallow CNN architecture (SRCNN) to learn the mapping function between bicubic-interpolated and HR image pairs, which demonstrates the effectiveness of CNN for image SR. Some methods~\cite{vdsr,drcn,drrn,memnet} follow the similar approach and employ deep networks with residual skip connections~\cite{vdsr,drrn}, or recursive supervision~\cite{drcn,drrn,memnet} for SISR and have achieved remarkable improvements. However,  these methods feed pre-interpolated LR image into networks to reconstruct a finer one with the same spatial resolution, which can increase the computational complexity for image SR. To solve this problem, other methods~\cite{espcn,lapsrn,edsr,rdn,dfrn} take original LR images as input and leverage transposed convolution~\cite{lapsrn,dfrn} or sub-pixel layer~\cite{edsr,espcn,rdn} to upscale final learned LR feature maps into HR space. Lim~\emph{et al.}~\cite{edsr} combine local residual skip connections and very deep network (EDSR) with wider convolution to further improve the image SR performance. 

The hierarchical features produced by immediate layers would provide useful information under different receptive fields for image restoration. Previous methods~\cite{lapsrn,drrn,edsr} fail to fully utilize hierarchical features and thus cause relatively-low SR performance. Tong~\emph{et al.}~\cite{srdensenet} adopt densely connected blocks~\cite{densenet} to exploit hierarchical features for HR image recovery. Zhang~\emph{et al.}~\cite{rdn} propose the residual dense block (RDB) and form a residual dense network (RDN) to extract abundant local features via dense connected convolutional layers and local residual learning. Nevertheless, almost all CNN based SISR methods simply adopt single-path feedforward architecture to enrich the feature representations from the input for the final prediction. By this way, the former low-level features is lacking incorporation with later high-level features. Thus later states cannot fuse the informative contextual information from previous states for discriminative feature representations. On the other hand, these methods introduce standard residual or dense connections to help the information flow propagation and alleviate the training difficulty, which can ignore the importance of different states within these connections.

In this paper, to address the problems mentioned above and mitigate the restricted multi-level context incorporation of solely feedforward architectures, we propose a novel deep interleave network (DIN) for image SR. The proposed DIN consists of multiple branches from the LR input to the predicted HR image, which learns how information at different states should be combined for image SR. Specifically, in each branch, we propose weighted residual dense block (WRDB) composed of cascading residual dense blocks to exploit hierarchical features that gives more clues for SR reconstruction. In the WRDB, we assign different weighted parameters to different inputs for more precise features aggregation and propagation, where the parameters can be optimized adaptively during training process. The WRDBs in adjacent interconnected branches interleave horizontally and vertically to progressively fuse the contextual information from different states. In this kind of design, the later branches can generate more powerful feature representations in combination with former branches. Besides, to improve the discriminative ability of our DIN for high-frequency details recovery, at each interleaved node among adjacent branches, we propose and attack the asymmetric co-attention (AsyCA) to adaptively emphasize the informative features from different states and generate trainable weights for feature fusion. After that, global feature fusion is further utilized in LR space for better HR images recovery.

Overall, the main contributions of our work are summarized as follows:
\begin{itemize}
\item We propose a novel deep interleaved network (DIN) which employs a multi-branch framework to fully exploit informative hierarchical features and learn how information at different states should be combined for image SR. Extensive experiments on public datasets demonstrate the superiority of our DIN over state-of-the-art methods.
\item We propose weighted residual dense block (WRDB) composed of multiple residual dense blocks in which different weighted parameters are assigned to different inputs for more precise features aggregation and propagation. The weighted parameters can be optimized adaptively during training process.
\item In our multi-branch DIN, the asymmetric co-attention (AsyCA) is proposed and attacked to the interleaved nodes to adaptively emphasize informative features from different states, which can generate trainable weights for feature fusion and further improve the discriminative ability of networks for high-frequency details recovery.
\end{itemize}

\section{Related Work}
SISR has recently achieved dramatic improvements using deep learning based methods. Dong~\emph{et al.}~\cite{srcnn} propose a 3-layer convolutional neural network (SRCNN) to minimize the mean square error between the bicubic-interpolated image and HR image for image SR, which significantly outperforms traditional sparse coding SR algorithms. Kim~\emph{et al.}~\cite{vdsr} construct a very deep SR network (VDSR) to exploit the contextual information spreading over large image regions. Motivated by the recursive supervision in DRCN~\cite{drcn} and residual learning in ResNet~\cite{resnet}, Tai~\emph{et al.}~\cite{drrn} introduce a deep recursive residual network (DRRN) that combines the recursive learning and residual connections to control the number of parameters while increasing the depth (52 layers) for image SR. Lim~\emph{et al.}~\cite{edsr} present a deeper and wider SR networks (EDSR) by cascading large number of residual blocks, which achieves dramatic performance and demonstrate the effectiveness of depth in image SR. 

Recently, many CNN-based SR methods employ different connections~\cite{srdensenet,memnet,carn,rdn,srfbn} in very deep networks to exploit hierarchical features for accurate image details recovery. Inspired by the densely skip connections in DenseNet~\cite{densenet}, Tong~\emph{et al.}~\cite{srdensenet} propose the densely connected SR network (SRDenseNet), which simply employs DenseNet as main architecture to improve information flow through the network for image SR. In~\cite{memnet}, a persistent memory network (MemNet) is introduced, which adopts the densely connected architecture in both local and global way to learn multi-level feature representations. Zhang \emph{et al.}~\cite{rdn} present the residual dense network (RDN) to exploit hierarchical features via the cascaded residual dense blocks (RDB). Li~\emph{et al.}~\cite{srfbn} propose an image
super-resolution feedback network (SRFBN) which combines the recurrent neural network (RNN) and feedback mechanism to refine low-level representations with high-level information. 

\begin{figure*}[t]
\centering
\includegraphics[width=1.64\columnwidth]{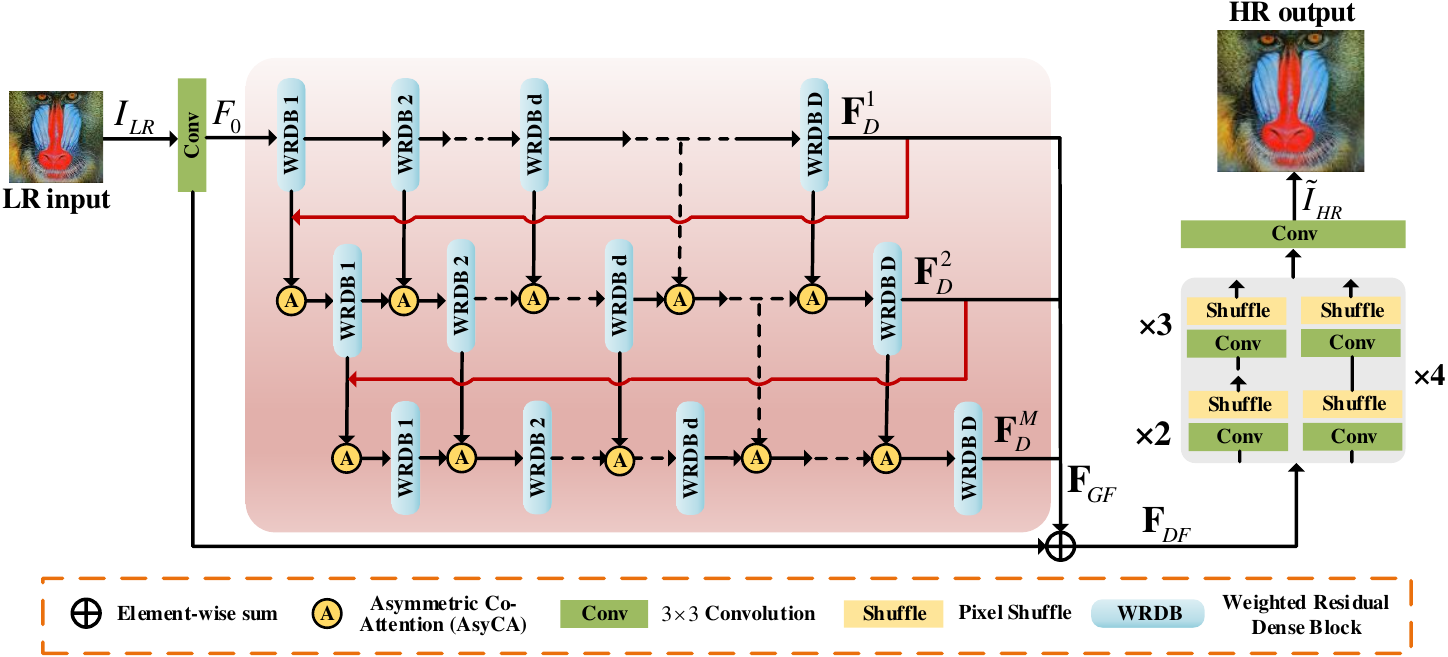} 
\caption{The architecture of our proposed deep interleaved network (DIN) for image SR.}
\label{fig1}
\end{figure*}

\section{Proposed Method}
In this section, we first elaborate the architecture of our deep interleaved network (DIN) for image SR in details and then suggest the interleaved multi-branch framework, and the asymmetric co-attention (AsyCA), which are the core of the proposed method.

\subsection{Network Architecture}
The proposed DIN, as illustrated in Fig. 1, the proposed DIN mainly contains four parts: shallow feature extraction network, the interleaved multi-branch framework composed of multiple cascading WRDBs in each branch, the asymmetric co-attention (AsyCA), and a upsampling reconstruction module.

Here, let's denote $I_{LR}$ as the LR input of our DIN and $I_{HR}$ is the corresponding HR image. As sketched in Fig. 1, one $3\times3$ convolutional layer is applied to extract the shallow feature $\mathbf{F}_0$ from the given LR input
\begin{equation}
\mathbf{F}_0 = H_0(I_{LR})
\label{eq1}
\end{equation}
where $H_{0}(\cdot)$ represents the convolution operation. $\mathbf{F}_0$ serves as the input fed into later multi-branch based feature interleaving and fusion, which produces deep feature as 
\begin{equation}
\mathbf{F}^M_D = H_{MBFI}(I_{LR})
\label{eq2}
\end{equation}
where $H_{MBFI}(\cdot)$ denotes the function of the multi-branch feature interleaving module (light red rectangle in Fig. 1), which consists of $M$ branches in which each branch contains $D$ WRDBs. $\mathbf{F}^M_D$ is the output of this module. Then we conduct global feature fusion (GFF) on the outputs from all $M$ branches. Thus we further have 
\begin{equation}
\mathbf{F}_{GF} = H_{GFF}(\mathbf{F}^1_D, \mathbf{F}^2_D,..., \mathbf{F}^M_D)
\label{eq3}
\end{equation}
where $\mathbf{F}_{GF}$ is the output feature of GFF by a composite function $H_{GFF}(\cdot)$. After that, long residual skip connection is introduced to stabilize the training of very deep network and can be represented as
\begin{equation}
\mathbf{F}_{DF} = \mathbf{F}_0 + \mathbf{F}_{GF}
\label{eq4}
\end{equation}
where $\mathbf{F}_{DF}$ is the output feature-maps by such residual learning. Finally, the output LR feature $\mathbf{F}_{DF}$ is upscaled via a upsampling reconstruction module to produce the HR image. In this work, we adopt the sub-pixel layer in ESPCN~\cite{espcn} with one convolutional layer for HR image reconstruction
\begin{equation}
\tilde{I}_{HR} = H_\uparrow(\mathbf{F}_{DF}) = H_{DIN}(I_{LR})
\label{eq5}
\end{equation}
where $\tilde{I}_{HR}$ and $H_\uparrow(\cdot)$ denote the generated HR image and corresponding upscale function respectively. $H_{DIN}(\cdot)$ represents the whole mapping function between $I_{LR}$ and $\tilde{I}_{HR}$.

We adopt $L_1$ loss to optimize the proposed network. Given a training dataset with $N$ LR images and their HR counterparts, denoted as $\left\{I^i_{LR}, I^i_{HR} \right\}^N_{i=1}$, the goal of training our DIN is to optimize the $L_1$ loss function:
\begin{equation}
L(\Theta) = \frac{1}{N}\sum_{i=1}^N||H_{DIN}(I^i_{LR}) - I^i_{HR}||_1
\label{eq6}
\end{equation}
where $\Theta$ denotes the learned parameter set of our proposed DIN. 

\subsection{Interleaved Multi-Branch Framework}
Our DIN follows a multi-branch pattern allowing multiple inter-connected branches to interleave and fuse at different states. The main point of our interleaved multi-branch framework lies in a progressively cross-branch feature interleaving and cascading WRDBs in each branch. In the first basic branch, supposing there are $D$ WRDBs, the output $\mathbf{F}^1_D$ of the first branch with $D$ WRDBs can be represented as 
\begin{equation}
\mathbf{F}^1_D = H^1_D(H^1_{D-1}(\cdot\cdot\cdot (H^1_1(\mathbf{F}_0))\cdot\cdot\cdot ))
\label{eq7}
\end{equation}
where $H^1_D(\cdot)$ denotes the operation of the $D^{th}$ WRDB. $H^1_D(\cdot)$ can be a composite function.

As shown in Fig. 1, we iteratively replicate the first basic branch multiple times, in which the sub-network at each branch can be regarded as a refinement process by continuously fusing the features from different states. For better description, we first denote the learning process of the basic single branch, as $\mathbf{F}_D = \Phi_D(\mathbf{F})$. In each branch, the output of the $(d-1)^{th}$ block $\Phi_{d-1}$ is the input of $d^{th}$ block $\Phi_{d}$. Thus the whole process of one branch can be formulated as
\begin{equation}
\mathbf{F}_D = \Phi_{d}(\Phi_{d-1}(...(\Phi_1(\mathbf{F}))...))
\label{eq8}
\end{equation}
Then, the features of these blocks in the same depth from adjacent branches respectively are fused to incorporate former low-level contextual information into current information flow for more powerful feature representations generation. The output of a certain block in previous branch is contributed to the input of its corresponding block in current branch. For the $m^{th}$ branch, the process of the $d^{th}$ block can be defined as $\Phi^m_d$. The block in previous branch is $\Phi^{m-1}_d$. Our multi-branch feature interleaving can be formulated as
\begin{equation}
\mathbf{F}^m_d=\left\{
             \begin{array}{lr}
             \Phi^m_d(S^m_d([\mathbf{F}^{m-1}_D, \mathbf{F}^{m-1}_1])), & \text{if}\; d=1 \\
             \Phi^m_d(S^m_d([\mathbf{F}^{m-1}_d, \mathbf{F}^m_{d-1}])), & otherwise.\\
             \end{array}
\right.
\label{eq9}
\end{equation}
where $S^m_d(\cdot)$ denotes the fusion operation at the interleaved nodes (orange circles in Fig. 1).  $\mathbf{F}^m_d$ is the output of the $d^{th}$ block in branch $m$. Note that $m$ and $d$ are integers in range $[1, M]$ and $[1, D]$, respectively.  

\begin{figure}[t]
\centering
\includegraphics[width=0.80\columnwidth]{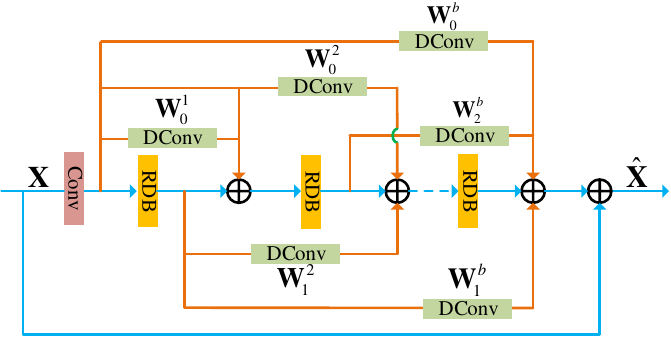}
\caption{Weighted residual dense block architecture, where ``DConv'' denotes the depth-wise convolution.}
\label{fig2}
\end{figure}

\subsubsection{Weighted Residual Dense Block}
The propose weighted residual dense block (WRDB) composed of cascading residual dense blocks (RDBs) to exploit hierarchical features that give more clues for SR reconstruction. In the WRDB, we assign different weighted parameters to different inputs for more precise features aggregation and propagation. The configuration of our constructed WRDB is depicted in Fig. 2. Supposing there are $b$ RDBs in one WRDB, given a input feature $\mathbf{X}\in\mathbb{R}^{C\times H \times W}$, where $H$ and $W$ are the spatial height and width of a feature map. $C$ is the number of input channels. The output of the $b^{th}$ RDB can be formulated as 
\begin{equation}
\begin{split}
\mathbf{X}_b&= sum(\mathbf{W}^b_0*\mathbf{X}_0, \mathbf{W}^{b}_1*\mathbf{X}_1,...,\mathbf{W}^b_{b-1}*\mathbf{X}_{b-1}) \\
\mathbf{X}_0 &= H_{0, d}(\mathbf{X})
\end{split}
\label{eq10}
\end{equation}
where $sum(\cdot)$ represents the element-wise sum operation. $H_{0, d}(\cdot)$ denotes the convolution operation of the first convolutional layer in the WRDB. $\mathbf{W}^b_{0}$ denotes the scaling weight set for the shortcut from the first convolutional layer to the $b^{th}$ RDB. $\mathbf{W}^b_{b-1}$ denotes the weight set for the connection from the $(b-1)^{th}$ block to the $b^{th}$ block. We employ depth-wise convolution to perform as the rescaling operation. Compared to the standard convolution, for the $C$ channels input $\mathbf{F}\in\mathbb{R}^{C\times H \times W}$, the depth-wise convolution applies a single filter on each input channel, which can be seen as assigning a weight parameter to each feature map, respectively. Besides, the computational cost of depth-wise convolution is
\begin{equation}
D_K \cdot D_K \cdot C \cdot H \cdot W
\label{eq11}
\end{equation}
where $D_K$ is the spatial dimension of the kernel. We can observe that the depth-wise convolutional layer with kernel size of $1\times1$ can involve very low computation and parameters. Therefore, we can easily employ depth-wise convolutional layers in our WRDB to rescale different inputs within the shortcuts and pass more detailed information flow across multiple states.

\subsection{Asymmetric Co-Attention}The aim of our proposed asymmetric co-attention (AsyCA) is to adaptively emphasize important information from different states at the interleaved nodes (yellow circles in Fig. 1) and generate trainable weights for feature fusion. The structure of AsyCA is illustrated in Fig. 3. Given features $\mathbf{X}_1$ and $\mathbf{X}_2$ are both with size of ${C\times H \times W}$. We first conduct concatenation on the two features
\begin{equation}
\widetilde{\mathbf{X}} = concat(\mathbf{X}_1, \mathbf{X}_2)
\end{equation}
\label{eq11}
where $concat(\cdot)$ denotes the concatenation operation. Then a $1\times1$ convolutional layers is used to integrate features coming from different branches. 
\begin{equation}
\mathbf{U} = H_{Int}(\widetilde{\mathbf{X}})
\label{eq12}
\end{equation}
where $H_{Int}(\cdot)$ indicates the integration function. We denote $\mathbf{U}=[\mathbf{u}_1, \mathbf{u}_2,\cdot\cdot\cdot, \mathbf{u}_C]$ as the output feature, which consists of $C$ feature maps with size of ${H \times W}$.

\begin{figure}[t]
\centering
\includegraphics[width=0.98\columnwidth]{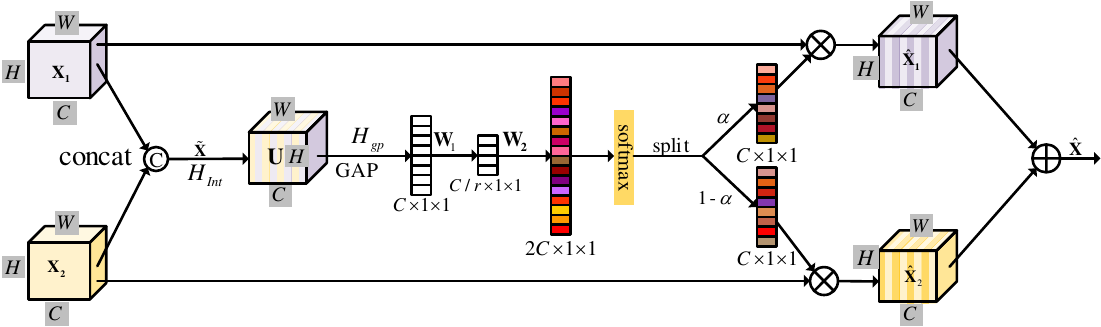}
\caption{The proposed asymmetric co-attention (AsyCA) architecture. ``GAP'' denotes the global average pooling}
\label{fig3}
\end{figure}

We then squeeze the global spatial information of $\mathbf{U}$ into a channel descriptor by a global average pooling to generate a channel-wise summary statistic $\mathbf{z}\in\mathbb{R}^{C\times1\times1}$. The $c^{th}$ element of $\mathbf{z}$ can be computed by shrinking $\mathbf{U}$ through spatial dimensions ${H \times W}$
\begin{equation}
z_c = H_{gp}(\mathbf{u}_c) = \frac{1}{H\times W}\sum^H_{i=1}\sum^{W}_{j=1}\mathbf{u}_c(i, j)
\label{eq13}
\end{equation}
where $\mathbf{u}_c(i, j)$ is the value at position $(i, j)$ of the $c^{th}$ channel $\mathbf{u}_c$. In order to fully capture channel-wise dependencies, we utilize a gating mechanism as SENet~\cite{senet} by forming a bottleneck with two $1\times1$ convolutional layers perform as dimensionality-reduction and -increasing with reduction ratio $r$
\begin{equation}
\mathbf{s} = \mathbf{W}_2*\sigma(\mathbf{W}_1*\mathbf{z})
\label{eq14}
\end{equation}
where $*$ denotes convolution operation and $\sigma(\cdot)$ represents the ReLU activation function. $\mathbf{W}_1\in\mathbb{R}^{C/r\times C \times H\times W}$ and $\mathbf{W}_2\in\mathbb{R}^{2C\times C/r\times H\times W}$ are the learned weights of the two convolutional layers. We obtain the final channel statistics $\mathbf{s}$ with size of $2C\times 1\times 1$. Then, we adopt a softmax operator to calculate the attention across channels and split the output into two chunks. This process can be formulated as 
\begin{equation}
\begin{split}
\alpha_c &= \frac{exp(\mathbf{V^c_1s})}{exp(\mathbf{V^c_1s})+exp(\mathbf{V^c_2s})}\\
1-\alpha_c &= \frac{exp(\mathbf{V^c_2s})}{exp(\mathbf{V^c_1s})+exp(\mathbf{V^c_2s})}
\end{split}
\label{eq15}
\end{equation}
where $\mathbf{V_1}\in\mathbb{R}^{C \times 1\times 1}$ and $\mathbf{V_2}\in\mathbb{R}^{C \times 1\times 1}$ denote the attention attention vector of $\mathbf{X_1}$ and $\mathbf{X_2}$, respectively. $\mathbf{V^c_1}$ is the $c^{th}$ row of $\mathbf{V_1}$ and $\alpha_c$ is the corresponding element of $\pmb{\alpha}$. The final feature $\widehat{\mathbf{X}}$ is obtained through the attention weights on the input features $\mathbf{X_1}$ and $\mathbf{X_2}$
\begin{equation}
\widehat{\mathbf{X}}_c = \alpha_c \cdot \mathbf{x}_{1, c} + (1 - \alpha_c) \cdot \mathbf{x}_{2, c}
\label{eq16}
\end{equation}
where $\widehat{\mathbf{X}}=[\widehat{\mathbf{x}}_1, \widehat{\mathbf{x}}_2, ..., \widehat{\mathbf{x}}_c]$ and $\widehat{\mathbf{x}}_c\in\mathbb{R}^{H\times W}$. $\mathbf{x}_{1,c}$ and $\mathbf{x}_{2,c}$ denote the $c^{th}$ elements of $\mathbf{X_1}$ and $\mathbf{X_2}$, respectively. By this way, our proposed AsyCA can adaptively adjust the important information from two adjacent branches and generate more discriminative feature representations.

\subsection{Implementation Details}
In our proposed DIN, We set the number of branches $M$ as 4. The initial shallow feature extraction layer have 64 filters with kernel size of $3\times3$. In each branch, we use $D=5$ WRDBs and each WRDB contains a $3\times3$ convolutional layer and 3 six-layer RDBs. The number of convolutional layers per RDB is 6, and the growth rate is 32. We set $3\times3$ as the size of all convolutional layers except that in AsyCA, whose kernel size is $1\times1$. The convolutional layers in each RDB has 64 filters followed by LeakyReLU with negative slope value 0.2. We utilize $1\times1$ depth-wise convolutional layer to conduct the densely weighted connections (DWCs) within each WRDB. 

\section{Experiments}
In this section, we conduct ablation study to investigate the effectiveness of each component in the proposed DIN. Then, we compared our models with other state-of-the-art image SR methods on pubic benchmark datasets.

\subsection{Setup}
\subsubsection{Datasets and Metrics}Following~\cite{edsr,rdn}, we use 800 HR images from DIV2K dataset~\cite{div2k} as our training set. All LR images are generated from HR images by using the Matlab function \emph{imresize} with the bicubic interpolation. For testing, we evaluate our SR results on five public standard benchmark datasets: Set5~\cite{set5}, Set14~\cite{set14}, BSD100~\cite{bsd100}, Urban100~\cite{selfexsr}, and Manga109~\cite{managa109}. All the SR results are evaluated with PSNR and SSIM on Y channel of the transformed YCbCr color space.

\subsubsection{Training Details}During training, we augment the training images by randomly flipping horizontally and rotating $90^{\circ}$. In each min-batch, 8 LR RGB patches with the size of $50\times50$ are randomly extracted as inputs. Our models are trained by Adam optimizer~\cite{adam} with $\beta_1=0.9$, $\beta_2=0.99$, and $\epsilon=10^{-8}$. The initial learning rate is set as $10^{-4}$ and then reduced to half every 200 epochs. We implement our networks with Pytorch framework on a Nvidia Titan Xp GPU.

\subsection{Ablation Study}
\subsubsection{Study of AsyCA and DWC.} In this subsection, we first investigate the effects on the key components in our proposed DIN, which contains the asymmetric co-attention (AsyCA) and the densely weighted connections (DWCs) in WRDB. Besides, we also investigate the effect of the global feature fusion (GFF) in our DIN. As shown in Table 1, the eight networks have the same structure. We first train a baseline model without these three components. We then add GFF to the baseline model. In the first and second columns, when both AsyCA and DWCs are removed, the PSNR on Set5 for $2\times$ SR is relatively low, no matter the GFF is used or not. After adding one of AsyCA or DWCs to the first models, we can validate that both AsyCA and DWCs can efficiently improve the performance of networks. It can be seen that the two components respectively combined with GFF perform better than only one component adding in the GFF model. This is because that our DWCs can assign different weight parameters on different for more precise information propagation and the AsyCA can emphasize important features from different states for more discriminative feature representations. When we use these three components simultaneously, the model (the last column) achieves the best performance. These quantitative comparisons demonstrate the effectiveness and benefits of our proposed AsyCA and DWCs.

\begin{table}[t]
    \scriptsize
    \caption{ Investigation of AsyCA, DWCs and GFF in our proposed DIN. We observe the best performance (PSNR) on Set5 with scaling factor $\times2$ in 50 epochs.}
    \label{tab1}
    \setlength{\tabcolsep}{1.8mm}
    \centering
    \begin{tabular}{|c|c|c|c|c|c|c|c|c|}
    \hline
         & \multicolumn{8}{c|}{Different combinations of AsyCA, DWCs and GFF} \\
         \hline
         \hline
         AsyCA & \XSolid & \XSolid & \Checkmark & \XSolid & \XSolid & \Checkmark & \Checkmark & \Checkmark \\ 
        \hline
        DWCs & \XSolid & \XSolid & \XSolid & \Checkmark & \Checkmark & \XSolid & \Checkmark & \Checkmark\\
        \hline  
        GFF & \XSolid & \Checkmark & \XSolid & \XSolid & \Checkmark & \Checkmark & \XSolid & \Checkmark\\
        \hline
        \hline  
        PSNR & 37.61 & 37.62 & 37.65 & 37.64 & 37.71 & 37.72 & 37.74 & \textbf{37.77}\\
        \hline
    \end{tabular} 
\end{table}

\subsubsection{Study of Feature Fusion Strategies}
Our proposed AsyCA can generate trainable weights for fusing the features from adjacent branches. In this subsection, we compare our method with another general feature fusion methods: concatenation (denote as concat) and element-wise sum (denote as sum). We visualize the convergence process of the three feature fusion strategies in Fig. 4. We can observe that the concatenation can achieve slightly higher PSNR performance than element-wise sum method but show serious fluctuation during training process. Compared to the concatenation and element-wise sum fusion, our proposed AsyCA can produce the best SR performance with more stable training process.

\begin{figure}[h]
\centering
\includegraphics[width=0.85\columnwidth]{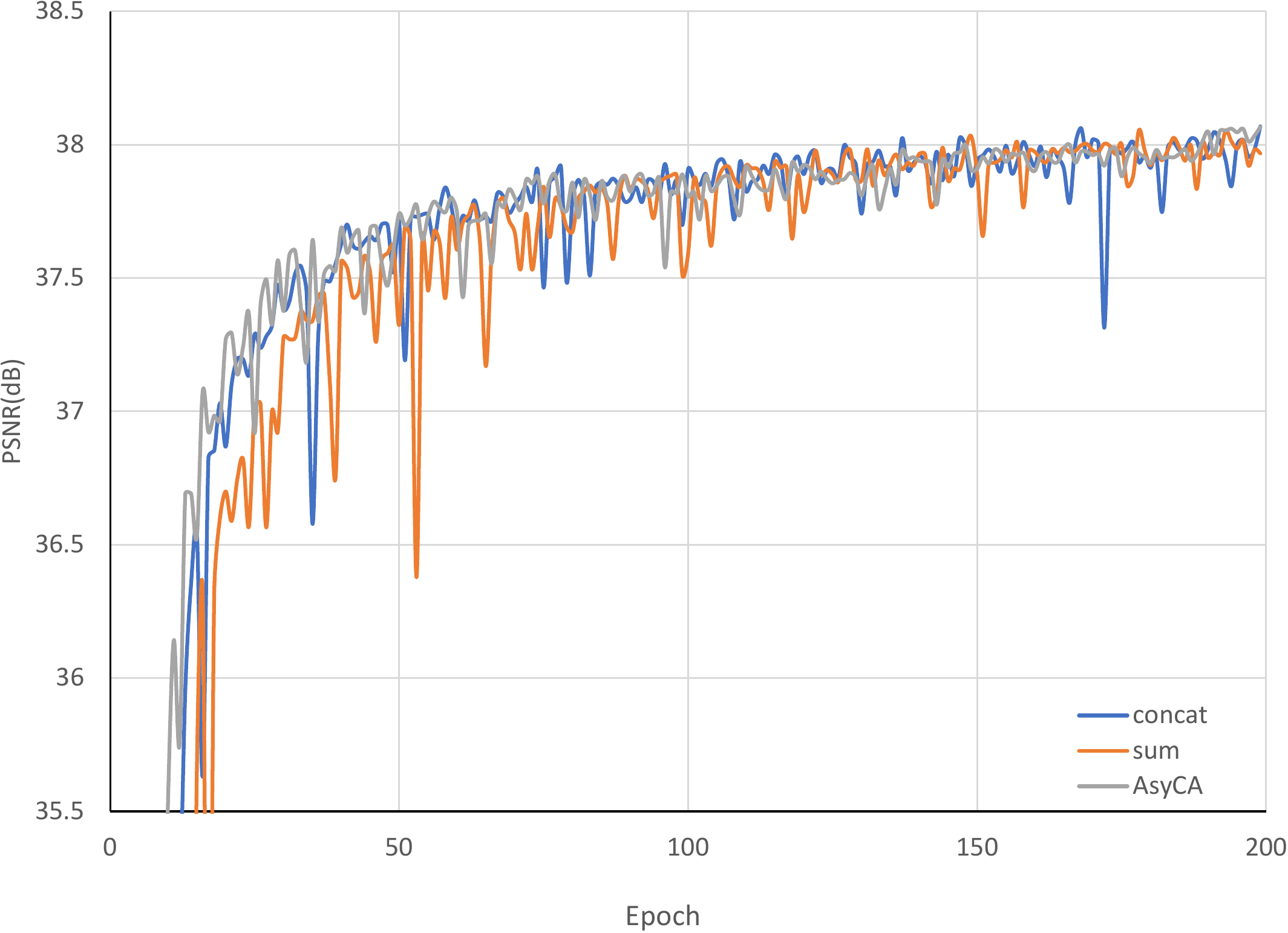} 
\caption{Convergence analysis of DIN with different feature fusion strategies: concatenation (concat), element-wise sum (sum), and AsyCA. }
\label{fig4}
\end{figure}

\begin{table*}[t]
    \scriptsize
    \caption{Benchmark results of different image SR methods. Average PSNR/SSIM values for scaling factor $\times2$, $\times3$, and $\times4$. The best performance is shown in {\color{red}{red}} and the
second best performance is shown in {\color{blue}{blue}}.}
    \label{tab2}
    \centering
    \begin{tabular}{|@{}c@{}|@{}c@{}|c|c|c|c|c|c|c|c|c|}
    \hline
	Dataset & Scale & Bicubic & SRCNN & LapSRN & DRRN & EDSR & SRFBN & RDN & DIN (ours) & DIN+ (ours)\\
    \hline
    \hline
    \multirow{3}{*}{Set5} & 2 & 33.66/0.9299 & 36.66/0.9542 & 37.52/0.9591 & 37.74/0.9591 & 38.11/0.9602 & 38.11/0.9609 & 38.24/0.9614 & {\color{blue}{38.26}}/{\color{blue}{0.9616}}& {\color{red}{38.29}}/{\color{red}{0.9617}}\\
	~ & 3 & 30.39/0.8682 & 32.75/0.9090 & 33.82/0.9227 & 34.03/0.9244 & 34.65/0.9280 & 34.70/0.9292 & 34.71/0.9296 & {\color{blue}{34.76}}/{\color{blue}{0.9298}} & {\color{red}{34.83}}/{\color{red}{0.9303}}\\
	~ & 4 & 28.42/0.8104 & 30.48/0.8628 & 31.54/0.8855 & 31.68/0.8888 & 32.46/0.8968 & 32.47/0.8983 & 32.47/0.8990 & {\color{blue}{32.67}}/{\color{blue}{.9006}} & {\color{red}{32.75}}/{\color{red}{0.9014}}\\
	\hline
	\hline
	\multirow{3}{*}{Set14} & 2 & 30.24/0.8688 & 32.45/0.9067 & 33.08/0.9130 & 33.23/0.9136 & 33.92/0.9195 & 33.82/0.9196 & 34.01/0.9212 & {\color{blue}{34.03}}/{\color{blue}{0.9214}} & {\color{red}{34.14}}/{\color{red}{0.9223}}\\
	~ & 3 & 27.55/0.7742 & 29.30/0.8215 & 29.79/0.8320 & 29.96/0.8349 & 30.52/0.8462 & 30.51/0.8461 & 30.57/0.8468 & {\color{blue}{30.65}}/{\color{blue}{0.8480}} & {\color{red}{30.72}}/{\color{red}{0.8491}}\\
	~ & 4 & 26.00/0.7027 & 27.50/0.7513 & 28.19/0.7720 & 28.21/0.7721 & 28.80/0.7876 & 28.81/0.7868 & 28.81/0.7871 & {\color{blue}{28.87}}/{\color{blue}{0.7890}} & {\color{red}{28.99}}/{\color{red}{0.7912}}\\
	\hline
	\hline
	\multirow{3}{*}{BSDS100} & 2 & 29.56/0.8431 & 31.36/0.8879 & 31.80/0.8950 & 32.05/0.8973 & 32.32/0.9013 & 32.29/0.9010 & 32.34/0.9017 & {\color{blue}{32.35}}/{\color{blue}{0.9018}} & {\color{red}{32.38}}/{\color{red}{0.9021}}\\
	~ & 3 & 27.21/0.7385 & 28.41/0.7863 & 28.82/0.7973 & 28.95/0.8004 & 29.25/0.8093 & 29.24/0.8084 & 29.26/0.8093 & {\color{blue}{29.29}}/{\color{blue}{0.8098}} & {\color{red}{29.33}}/{\color{red}{0.8107}}\\
	~ & 4 & 25.96/0.6675 & 26.90/0.7101 & 27.32/0.7280 & 27.38/0.7284 & 27.71/0.7420 & 27.72/0.7409 & 27.72/0.7419 & {\color{blue}{27.78}}/{\color{blue}{0.7437}} & {\color{red}{27.87}}/{\color{red}{0.7459}}\\
	\hline
	\hline
	\multirow{3}{*}{Urban100} & 2 & 26.88/0.8403 & 29.50/0.8946 & 30.41/0.9101 & 31.23/0.9188 & 32.93/0.9351 & 32.62/0.9328 & 32.89/0.9353 & {\color{blue}{33.11}}/{\color{blue}{0.9371}} & {\color{red}{33.27}}/{\color{red}{0.9383}}\\
	~ & 3 & 24.46/0.7349 & 26.24/0.7989 & 27.07/0.8272 & 27.56/0.8376 & 28.80/0.8653 & 28.73/0.8641 & 28.80/0.8653 & {\color{blue}{28.94}}/{\color{blue}{0.8682}} & {\color{red}{29.09}}/{\color{red}{0.8705}}\\
	~ & 4 & 23.14/0.6577 & 24.52/0.7221 & 25.21/0.7553 & 25.44/0.7638 & 26.64/0.8033 & 26.60/0.8015 & 26.61/0.8028 & {\color{blue}{26.85}}/{\color{blue}{0.8089}} & {\color{red}{27.13}}/{\color{red}{0.8144}}\\
	\hline
	\hline
	\multirow{3}{*}{Manga109} & 2 & 30.80/0.9339 & 35.60/0.9663 & 37.27/0.9740 & 37.60/0.9736 & 39.10/0.9773 & 39.08/0.9779 & 39.18/0.9780 & {\color{blue}{39.39}}/{\color{blue}{0.9785}} & {\color{red}{39.53}}/{\color{red}{0.9788}}\\
	~ & 3 & 26.95/0.8556 & 30.48/0.9117 & 32.19/0.9334 & 32.42/0.9359 & 34.17/0.9476 & 34.18/0.9481 & 34.13/0.9484 & {\color{blue}{34.46}}/{\color{blue}{0.9496}} & {\color{red}{34.68}}/{\color{red}{0.9507}}\\
	~ & 4 & 24.89/0.7866 & 27.58/0.8555 & 29.09/0.8893 & 29.18/0.8914 & 31.02/0.9148 & 31.15/0.9160 & 31.00/0.9151 & {\color{blue}{31.23}}/{\color{blue}{0.9173}} & {\color{red}{31.66}}/{\color{red}{0.9221}}\\
	\hline
\end{tabular}
\end{table*}

\subsection{Comparing with the state-of-the-arts}
\subsubsection{Quantitative Comparison}We compare our DIN with 7 state-of-the-art image SR methods: SRCNN, LapSRN, DRRN, MemNet, EDSR, SRFBN, and RDN. Self-ensemble strategy~\cite{edsr} is utilized to further improve our DIN and we denote the self-ensembled DIN as DIN+. Table 2 shows quantitative comparisons for $2\times$, $3\times$, and $4\times$ image SR. Obviously, compared with other methods, our DIN+ performs the best results on all the datasets on various scaling factors. Besides, our DIN outperforms all methods in terms of both PSNR and SSIM on all datasets, especially the most state-of-the-art method RDN that involves more parameters than ours. Even without the self-ensemble strategy, the proposed DIN achieves better SR performance compared with other image SR methods. 

\begin{figure}[t]
\centering
\includegraphics[width=1.00\columnwidth]{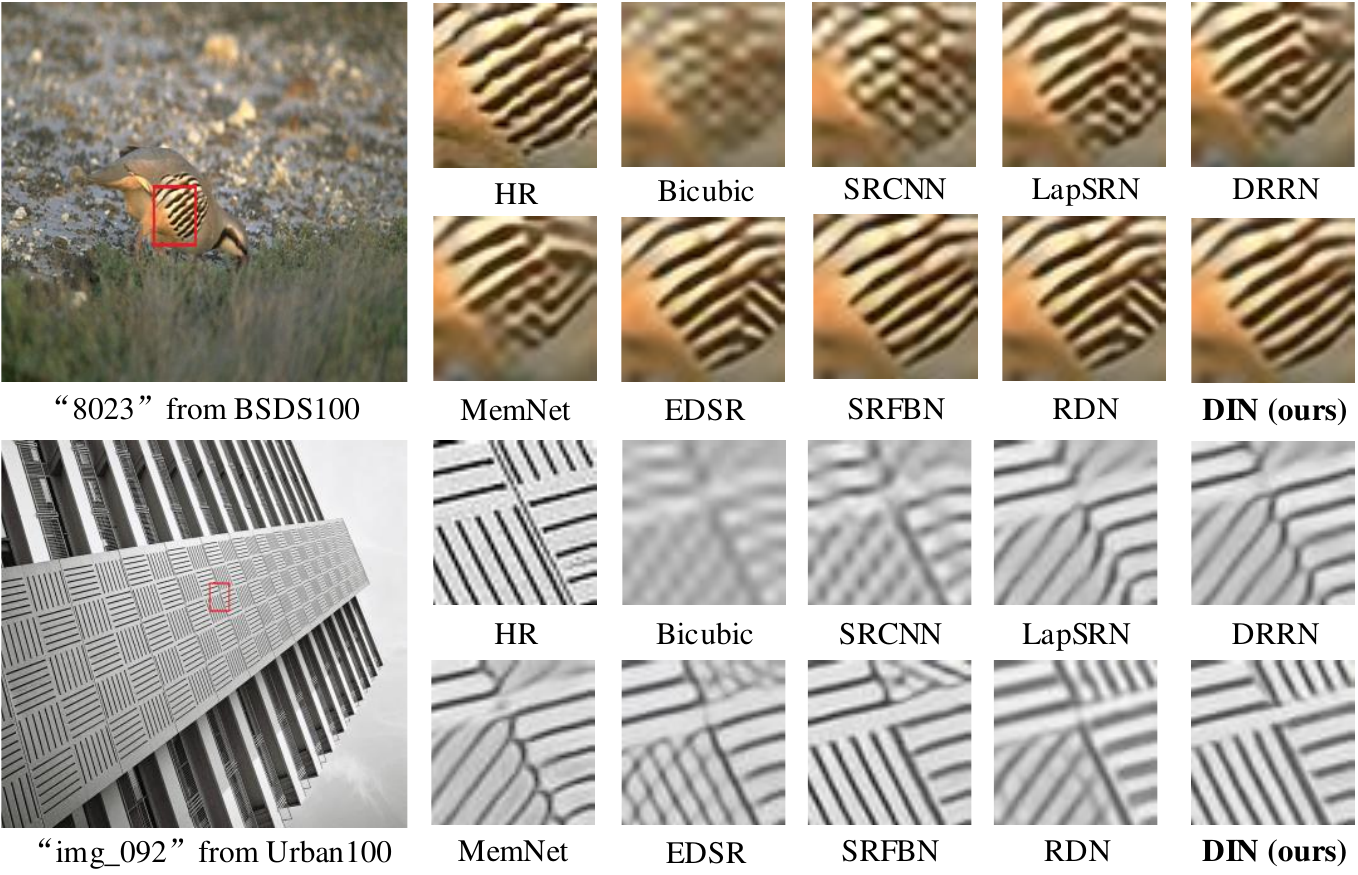} 
\caption{Visual comparison for $4\times$ SR on BSDS100 and Urban100 datasets.}
\label{fig5}
\end{figure}

\subsubsection{Visual Quality}We show SR results with scaling factor $\times4$ in Fig. 5. In general, the proposed DIN can yield more convincing results. For the SR results of the ``8023'' from BSDS100, most methods tend to produce HR image with heavy blurry contents. In contrast, our DIN obtains the SR results with clearer contour and much fewer blurs. For the ``img\_092'' from Urban100, the results of EDSR and RDN suffer from wrong texture directions and serious artifacts. Though SRFBN alleviates it to a certain extent, the generated HR still contains some misleading contents. Our DIN can recover clearer details and reliable textures which are more faithful to the ground truth.

\subsubsection{Model size}
Table 3 shows the performance and model size of recent very deep CNN-based image SR models. Among these models, MemNet and SRFBN contain much fewer parameters at the cost of performance degradation. Our proposed DIN can obtain superior performance than EDSR and RDN but has fewer parameters, which demonstrates that our DIN can achieve a good trade-off between SR performance and model complexity.

\section{Conclusion}
In this paper, we propose a novel deep interleaved network (DIN) to reconstruct the HR image from a given LR image by employing a multi-branch framework to interleave and fuse at different states. Specifically, in each branch, we propose weighted residual dense block (WRDB) to exploit hierarchical features, which assigns different weighted parameters to different inputs for more precise features aggregation and propagation. The WRDBs in adjacent interconnected branches interleave horizontally and vertically to progressively fuse the contextual information from different states. In addition, at each interleaved node among adjacent branches, we propose and attack the asymmetric co-attention (AsyCA) to adaptively emphasize the informative features from different states and generate trainable weights for feature fusion, which can improve the discriminative ability of our network for high-frequency details recovery. Comprehensive experimental results demonstrate that our DIN achieves superiority over the state-of-the-art image SR methods.
\begin{table}[t]
    \scriptsize
    \caption{Parameter number (Param.), and PSNR (dB) comparisons. The PSNR values are based on Set14 with scaling factor $\times2$.}
    \label{tab3}
    \setlength{\tabcolsep}{1.8mm}
    \centering
    \begin{tabular}{|@{}c@{}|c|c|c|c|c|c|c|c|}
    \hline
          Methods & LapSRN & DRRN & MemNet & EDSR & SRFBN & RDN & \textbf{DIN}\\
         \hline
         \hline
         Param. & 812K & 297K & 677K & 43M & 3.63M & 22M & 19.88M \\ 
        \hline
        PSNR & 33.08 & 33.23 & 33.28 & 33.92 & 33.82 & 34.01 & \textbf{34.03} \\
        \hline  
    \end{tabular} 
\end{table}

\section{Acknowledgments}
This work was supported in part by the Fundamental Research Funds for the Central Universities (2019JBZ102) and the National Natural Science Foundation of China (No. 61972023).

\bibliographystyle{named}
\bibliography{ijcai20}
\end{document}